\documentclass[letterpaper, 10 pt, conference]{ieeeconf}  

\IEEEoverridecommandlockouts                              

\usepackage{graphics} 
\usepackage{epsfig} 
\usepackage{mathptmx} 
\usepackage{times} 
\usepackage{amsmath} 
\usepackage{amssymb}  
\usepackage[utf8]{inputenc} 
\usepackage[T1]{fontenc}    
\usepackage{url}            
\usepackage{booktabs}       
\usepackage{amsfonts}       
\usepackage{nicefrac}       
\usepackage{microtype}      
\usepackage{subcaption}
\usepackage{siunitx}
\usepackage{hyperref}
\usepackage{amsmath,amssymb,amsfonts}
\usepackage{algorithmic}
\usepackage{graphicx}
\usepackage{textcomp}
\usepackage{footnote}
\usepackage{enumerate}
\usepackage{adjustbox}
\usepackage{dblfloatfix}
\usepackage{multirow}
\usepackage[bottom]{footmisc} 
\usepackage{tablefootnote}

\usepackage[table]{xcolor}
\definecolor{blue}{rgb}{0.00,0.00,1.00}
\definecolor{green}{rgb}{0.30, 0.50,0.00}
\definecolor{lightgray}{gray}{0.9}
\definecolor{purple}{rgb}{0.60,0.10,0.90}

\usepackage{biblatex}
\title{\LARGE \bf
Early or Late Fusion Matters: Efficient RGB-D Fusion in Vision Transformers for 3D Object Recognition
}

\author{Georgios Tziafas$^{1}$ and Hamidreza Kasaei$^{1}$
\thanks{$^{1}$Department of Artificial Intelligence,
        University of Groningen, The Netherlands
        {\tt\small \{g.t.tziafas,hamidreza.kasaei\}@rug.nl}}}

\begin{document}

\maketitle
\thispagestyle{empty}
\pagestyle{empty}

\begin{abstract}
The Vision Transformer (ViT) architecture has established its place in computer vision literature, however, training ViTs for RGB-D object recognition remains an understudied topic, viewed in recent literature only through the lens of multi-task pretraining in multiple vision modalities.
Such approaches are often computationally intensive, relying on the scale of multiple pretraining datasets to align RGB with 3D information.
In this work, we propose a simple yet strong recipe for transferring pretrained ViTs in RGB-D domains for 3D object recognition, focusing on fusing RGB and depth representations encoded jointly by the ViT.
Compared to previous works in multimodal Transformers, the key challenge here is to use the attested flexibility of ViTs to capture cross-modal interactions at the downstream and not the pretraining stage.
We explore which depth representation is better in terms of resulting accuracy and compare early and late fusion techniques for aligning the RGB and depth modalities  within the ViT architecture.
Experimental results in the Washington RGB-D Objects dataset (ROD) demonstrate that in such RGB $\rightarrow$ RGB-D scenarios, late fusion techniques work better than most popularly employed early fusion. 
With our transfer baseline, fusion ViTs score up to 95.4\% top-1 accuracy in ROD, achieving new state-of-the-art results in this benchmark.
We further show the benefits of using our multimodal fusion baseline over unimodal feature extractors in a synthetic-to-real visual adaptation as well as in an open-ended lifelong learning scenario in the ROD benchmark, where our model outperforms previous works by a margin of >8\%.
Finally, we integrate our method with a robot framework and demonstrate how it can serve as a perception utility in an interactive robot learning scenario, both in simulation and with a real robot.
\end{abstract}

\section{Introduction}
Transfer learning approaches for computer vision have a long-standing tradition for image classification, most popularly using Convolutional Neural Networks (CNNs).
More recently, the Vision Transformer (ViT) \cite{vit} architecture and its variants \cite{swin, DeiT, BEiT} have shown promising transfer results, providing flexible representations that can be fine-tuned for downstream tasks, also in few-shot settings \cite{fewshotvit}.
This capability however comes at the cost of data inefficiency \cite{vitsurvey}, as performance gains over CNNs are noticed in Transformers that are pretrained in large-scale datasets, such as ImageNet21k \cite{imagenet21k} and JFT-300M \cite{jft300m}. 
When moving from RGB-only to view-based 3D object recognition (RGB-D), a dataset of similar magnitude for pretraining is amiss, granting RGB-D representation learning a topic that has yet to grow.
Recent alternative directions include transferring from models pretrained on collections of multimodal datasets \cite{omnivore, CMX, OMNIMAE}, although they focus on scene-level tasks, they are constrained to the use of the early fusion strategy and are often computationally intensive to fine-tune.

\begin{figure*}[t]
    \centering
    \includegraphics[width=1\textwidth]{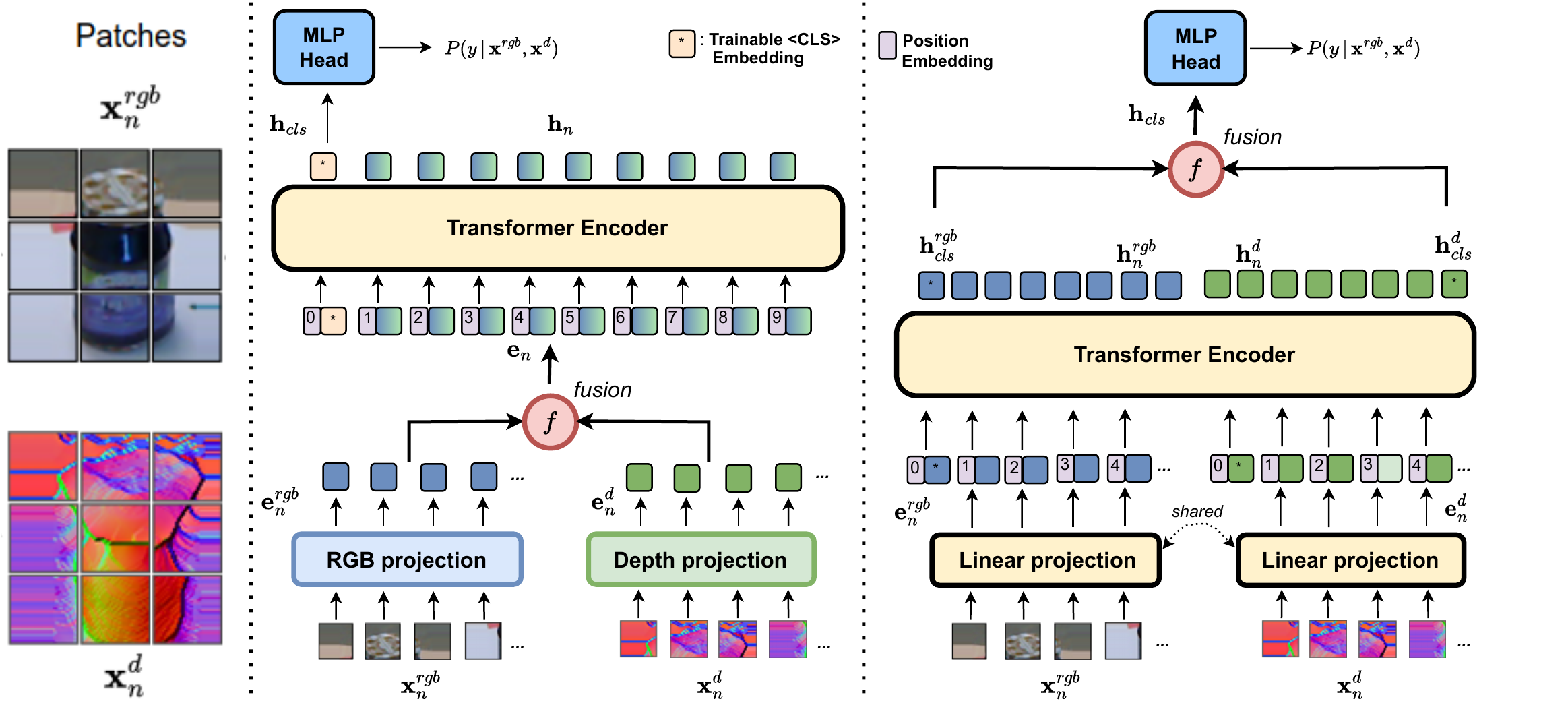}
    \caption{ \footnotesize{Two different baselines for fusing RGB-D representations in the ViT architecture. In \textit{early} fusion (\textit{left}), a separate projection is used for RGB and depth and the fused embeddings are fed to the encoder, providing a single \texttt{<CLS>} token. In \textit{late} fusion (\textit{right}), the same weights are used for projecting RGB and depth and the two modalities are fed separately to the encoder. The two final \texttt{<CLS>} tokens are fused to provide the final representation for classification.}}
    \label{fig:fig1}
\end{figure*}

In this work, we aim to address such limitations by revisiting the RGB-D object recognition task and study recipes for transferring an RGB-only pretrained ViT (i.e. in ImageNet1k \cite{deng2009imagenet}) into an RGB-D object-level dataset.
We begin by exploring different representation formats for the depth modality and design two variants that adapt ViT to fuse RGB and depth (see Fig.~\ref{fig:fig1}), namely: a) \textit{Early} fusion, where RGB and depth are fused before the encoder and RGB-D patches are represented jointly in the sequence, and b) \textit{Late} fusion, where we move the fusion operation after the encoder, leaving the patch embedders intact from their pretraining.
Our hypothesis is that when fine-tuning in small (or moderate) sized datasets, the late fusion baseline is very likely to perform better, as it doesn't change the representation of the input compared to the pretraining stage, but casts the challenge as a distribution shift in the input images (i.e. both RGB and depth are processed by the same weights and must be mapped to the same label).

Experimental results with the Washington RGB-D Objects dataset \cite{wrgbd} positively reinforce our hypothesis, as the late fusion baseline far outperforms the early variant. 
More interestingly, we show that with our late fusion recipe, ViTs achieve new state-of-the-art results in this benchmark, surpassing a plethora of methods that specifically study RGB-D fusion techniques for object recognition.
We conduct additional experiments to further demonstrate the representational strength of our approach in: a) a synthetic-to-real transfer scenario, where we show that with late fusion a synthetically pretrained ViT can surpass the performance of training in real data with only a few fine-tuning examples, and b) an open-ended lifelong learning scenario, where we show that our late fusion encoder outperforms unimodal versions of the same scale, even without fine-tuning, while outperforming previous works by a significant margin.
Finally, we demonstrate the applicability of our approach in the robotics domain by integrating our method with a simulated and real robot framework. In particular, we illustrate how the robot can be taught by a human user to recognize new objects and perform a table-cleaning task.
In summary, our main contributions are:
\begin{itemize}
    \item We experimentally find that late fusion performs better than early fusion in RGB $\rightarrow$ RGB-D transfer scenarios.
    \item We achieve new state-of-the-art results for RGB-D object recognition in the ROD \cite{wrgbd} benchmark.
    \item We show that our method can aid in SynROD  $\rightarrow$ ROD \cite{SynROD} few-shot visual domain adaptation scenario.
    \item We show that our method can be applied in an online lifelong robot learning setup, including experimental comparisons with previous works as well as simulation and real robot demonstrations. 
\end{itemize} 
        

\section{Related Works}
In this section, we discuss previous works on RGB-D fusion with CNNs for view-based object recognition, multimodal Transformers, as well as open-ended lifelong learning, which we include as an evaluation scenario in our experiments.

\subsection{RGB-D Fusion with CNNs}
As in RGB image classification, multiple traditional CNN-based approaches have replaced conventional approaches \cite{related11, related12} for extending to the RGB-D modalities.
The focus of such works lies in RGB-D fusion, where deep features extracted from CNNs are fused through a multimodal fusion layer \cite{related13} or custom networks \cite{related14}.
Rahman et. al. \cite{related15} propose a parallel three-stream CNN which processes two depth encodings in two streams and RGB in the last one.  
Cheng et. al. \cite{related16} proposed to integrate Gaussian mixture models with CNNs through fisher kernel encodings.
Zia et. al. \cite{related17} propose mixed 2D/3D CNNs which are initialized with 2D pretrained weights and extend to 3D to also incorporate depth. 
Such methods study how to inject fusion in the locally-aware CNN architecture. 
In contrary, in our work, we implement fusion as a pooling operation on multimodal Transformer embeddings and opt to gain cross-modal alignment via transferring from pretrained models.

\subsection{Multimodal Learning with Transformers}
In the absence of a large-scale RGB-D dataset for pretraining, recent works try to alleviate this bottleneck by pretraining on collections of datasets from multiple modalities \cite{omnivore, CMX, OMNIMAE} and rely on the flexibility of Transformers to capture cross-modal interactions.
However, such methods focus on scene/action recognition or semantic segmentation tasks, leaving the RGB-D object recognition task unexplored. 
Furthermore, they employ an early fusion technique for converting heterogeneous modalities (i.e., image, video) in the same sequence representation, leaving open questions of whether this is the best fusion technique in homogeneous modalities such as RGB-D, as well as if its the best fusion technique for directly transferring from one homogeneous modality to another, without the pretraining step.
Finally, they rely heavily on model capacity and specialized Transformer architecture variants (e.g Swin \cite{swin}) in order to enable multimodal pretraining to boost performance in unimodal downstream tasks. 
Such models set a high computational resource entry point for practitioners, casting them not widely accessible for fine-tuning in arbitrary datasets.

\subsection{Open-Ended Lifelong Learning} An emerging topic in deep learning literature, most commonly referred to as \textit{Lifelong} or \textit{Continual Learning}, studies the scenario of a learning agent continuously incorporating new experiences from an online data stream.
In the context of image classification, the challenge is stated as learning to classify images from an ever-shifting distribution, while avoiding the effect of catastrophic forgetting \cite{CLsurvey, A-GEM, iCaRL, CI1}.
Even though works for using Transformers in lifelong learning are starting to grow \cite{ll2, ll3}, to the best of our knowledge, this is the first work that touches on lifelong learning with Transformers for RGB-D object recognition.
However, we highlight that the focus of this work is not on lifelong learning algorithms, but rather to establish a baseline in the ROD benchmark for future references.

To that end, we adopt the interactive lifelong learning protocol proposed by previous works \cite{Kasaei2019OrthographicNetAD, kasaei2018towards, LocalLDA, GOOD}, where a human user introduces novel objects to the agent through human-robot interaction.
In this setup, GOOD \cite{GOOD} was first proposed as a global hand-crafted object descriptor that encodes a 3D object as a histogram, showing a good balance between accuracy, computation time, and robustness to noise. This descriptor is particularly suitable for robotic applications with limited resources. Kasaei et al. \cite{LocalLDA} extended Latent Dirichlet Allocation (LDA) \cite{Hoffman2010OnlineLF} to Local-LDA, which is a Bayesian method for learning representations for each object category incrementally and independently. They showed the application of Local-LDA in the context of open-ended 3D object category learning and recognition. OrthographicNet \cite{Kasaei2019OrthographicNetAD} first projects the 3D object into three orthographic views and then feeds them to a pretrained CNN network to obtain a feature vector for each projection. 
The obtained features are then fused using a pooling function to form a global description for the given 3D object.
In this work, we use a similar setup as OrthographicNet but pass single RGB-D views from a multimodal ViT instead of a CNN encoder.
        
\section{Approach}
Our goal is to have a single model that can be transferred to RGB-D downstream tasks while being pretrained solely in RGB. 
Even though the two modalities are homogeneous, the corresponding pixel distributions have a significant gap and different depth representations might aid in aligning them more closely.
To explore such possibilities, we experiment with several different representations of the depth image.
Unlike standard fine-tuning strategies, that try to model a new distribution under the same modality, in this work we wish to enable ViT to also learn from the depth modality, as well as learn how to successfully model the correspondences between the two modalities. 
To that end, we explore two different RGB-D representation fusion techniques.

\subsection{ViT Prerequisites}
The ViT model handles the visual input as a sequence of image patches. The original $H \times W$ image $\mathcal{I}$ is split into patches of size $h \times w$, resulting in a total of $N = \frac{H \cdot W}{h \cdot w}$ patches. 
Each patch is flattened into a single vector representation $\mathbf{x}_n \in \mathbb{R}^{3*h*w}$ and projected into an embedding space through a linear map $\mathcal{E}(\mathbf{x}_n)=\mathbf{e}_n, \; \mathcal{E}: \mathbb{R}^{3*h*w} \rightarrow \mathbb{R}^{D}$.
A trainable image-level embedding $\mathbf{e}_0$ (i.e. the \texttt{<CLS>} token representation) is stacked with the embeddings sequence and the patch embeddings are further added with positional encodings $\mathbf{p}_n$, either learned jointly or hand-crafted (e.g 2D sinusoid). 
The resulting sequence $\left [ \mathbf{e}_0, \{\mathbf{e}_n+\mathbf{p}_n \right\}_{n=1}^N ]$ is passed through $L$ layers of Transformer encoder blocks, resulting in the sequence of hidden states $[\mathbf{h}_n^l]_{n=0}^N, \; l=1,..,L $.
For downstream classification tasks, the final hidden \texttt{<CLS>} state $\mathbf{h}_{cls} \doteq  \mathbf{h}_0^L$ is fed into an MLP head, whose output is supervised using a standard cross-entropy loss over the task's target classes.

\subsection{Depth Representation}
\label{depth_rep}
In order to make ViT compatible with the RGB-D modality, we need to express the input depth map with the same format as RGB. 
In the standard ViT pipeline, the input RGB image is first resized to a fixed resolution (in base configuration, $H=W=224$), center-cropped, and then normalized according to the mean and standard deviation of the training dataset.
We experiment with three different types of depth representations, namely:
\begin{enumerate}
    \item \textbf{Raw depth} maps, truncated to a pre-set maximum depth value (e.g. around $3.5$ meters for Kinect) and clipped to $[0,255]$ range. We stack three instances of the resulting map to "convert" it to RGB.
    \item \textbf{HHA} transformations of the raw depth maps, which have shown to encode geometric properties, such as geocentric pose. To compute the transform, depth is first converted to a disparity map using the camera intrinsics. The HHA is then built as an image with three channels at each pixel, including horizontal disparity, height above ground, and the angle at the pixel’s local surface normal.
    \item \textbf{Surface Normal} reconstructions, which have shown to encode fine-grained 3D details about shape, texture and surface.
    These images are generated by and estimating surface normals at each point of the corresponding point cloud. The 3D vectors are back-projected to the camera reference frame and colorized separately in a channel.
\end{enumerate}

The resulting colorized depth image is fed into the same resize-crop-normalization preprocessing as in RGB.

\subsection{RGB-D Fusion Techniques}
We explore two different types of RGB-D fusion, aiming to determine which is the most accurate way to adapt pretrained RGB Transformers for RGB-D recognition tasks in the absence of large-scale RGB-D datasets. 
\paragraph{\textbf{Early Fusion}} In early fusion, the RGB-D representations are fused before the Transformer encoder, and the encoder is fine-tuned as-is in the multi-modal representations. Following \cite{omnivore}, we use a separate patch embedding layer for each modality, $\mathcal{E}^{rgb}$ and $\mathcal{E}^d$ and fuse the two representations before adding the position embeddings, using addition and L2 normalization:
\begin{equation}
    \mathbf{e_n}=\frac{\mathcal{E}^{rgb}(x_n^{rgb}) + \mathcal{E}^{d}(x_n^{d})}{\left \| \mathcal{E}^{rgb}(x_n^{rgb}) + \mathcal{E}^{d}(x_n^{d}) \right \|_2}
\end{equation}
We call this baseline the \textit{dual}-embedder, as it separates embeddings for the two modalities.
In our experiments we also implement the early fusion baseline with a \textit{joint-}embedder, stacking the two modalities channel-wise and using a single projection to embed them jointly.
In this architecture, a single \texttt{<CLS>} embedding is learned for the entire RGB-D pair. 
The key insight is that through the self-attention operation the joint RGB-D embedding $\mathbf{h}_{cls}$ will adapt to model the inter-modal alignment between the fused representations.
To make the pretrained ViT checkpoint compatible with the adapted architecture, we copy the weights of the pretrained patch embedder in both RGB and depth embedders.

\paragraph{\textbf{Late Fusion}} In late fusion, we pass the two images to the ViT encoder separately and aggregate their final \texttt{<CLS>} embeddings, before passing it to the classifier. We experiment with different types of late fusion operations $f: \mathbb{R}^D \times \mathbb{R}^D \rightarrow \mathbb{R}^{D'}$, such as max pooling ($D'=D$), averaging ($D'=D$) and concatenation ($D'=2 \cdot D$). 
In this baseline, the encoder has to learn how to classify both images of the same object view, while processing the two modalities separately. 
The final hidden representation fed to the head is the fusion of the two hidden states: $\mathbf{h}_{cls}=f(\mathbf{h}_{cls}^{rgb},\mathbf{h}_{cls}^{d})$

\subsection{Implementation Details}
In order to fine-tune the adapted ViT, we use a dataset-specific linear layer on top of the final \texttt{<CLS>} embedding with a softmax loss over the datasets' categorical distribution of labels.
For the early fusion baseline, we replicate the implementation by \cite{omnivore} and develop the RGB and depth embedding layers as 2D convolution layers with $D$ feature maps and kernel size and stride of $N$. 
The input channels are $3$ for RGB and depth separately in the dual version and $6$ for the joint embedder variant.
For training the late fusion baseline, we generate separate RGB and depth batches and interleave the latter in between the first, so that the model processes the RGB-D pair in pairs of two, even in the case of distributed parallel training with multiple GPUs/nodes.
We experiment with the default public configurations of ViT-{$x$}, where $x=$\{T,S,B,L\} for \{\textit{tiny}, \textit{small}, \textit{base}, \textit{large}\}.

\section{Experiments}
This section describes our evaluation setup and presents our results. 
It is organized as follows: First (Sec.~\ref{datasets}), we give the specifics of the RGB-D datasets used for training and the evaluation scheme. 
Then (Sec.~\ref{ablations}), we perform ablation studies for different variants of the depth representation and the fusion approach. 
We compare our best model with previous state-of-the-art in ROD and benchmark other RGB-D datasets (Sec.~\ref{wrgbd}).
We further explore the synthetic-to-real domain adaptation capabilities of our model in Sec.~\ref{synthetic-to-real}.
Finally, (Sec.~\ref{lifelong}), we study the performance of our approach in an online lifelong learning setup and (Sec.~\ref{robot}) demonstrate an interactive robot learning scenario after integrating our method with a robot framework.

\subsection{Datasets and Evaluation}
\label{datasets}

\textbf{Washington RGB-D Objects (ROD)} \cite{wrgbd} is a well-established benchmark for object recognition tasks in RGB-D domains. 
It contains up to $41 \,  877$ views from $300$ object instances, organized into $51$ categories, including common household objects (cups, bowls, mugs, etc). 
For depth representations, we use the surface normals as extracted from \cite{sota10}. 
Regarding evaluation, we perform the suggested experiment as in the original paper \cite{wrgbd}. 
In particular, we perform 10 trials, in each of which one instance per category is used for testing (total $\sim 7$k RGB-D pairs) and the rest for training (total $\sim 35$k RGB-D pairs).


\textbf{ARID40k} is a subset of a ARID \cite{arid}, containing $6000$ scenes rom $153$ different object instances in $51$ categories. 
We use the subset provided by the authors, containing $40\,713$ RGB-D crop pairs.
This dataset includes the ROD object distribution designed for evaluation in robotic-specific applications.

\textbf{Homebreweddb (HB)} is a scenes dataset \cite{HomebrewedDBRD} used for 6D pose estimation that features high-quality 3D-reconstructed models for a total of 33 instances. We repurpose for the recognition task by cropping RGB-D objects from validation sequences, for a total of $22\,935$ RGB-D samples.
Depth uses surface normal format.
This is a more challenging task, as it includes a more diverse distribution of objects (i.e., toys, household, industrial).

\textbf{OCID} \cite{OCID} comprises of 96 cluttered scenes, using $89$ object instances from ARID, as well as the YCB dataset \cite{YCB}.
We use the subset of RGB-D crops provided by the authors, for a total of $5\,735$ samples.
The cluttered nature of scenes in this dataset makes the recognition task interesting due to noisy crops caused by occlusion.

\subsection{Ablation Studies}
\label{ablations}

\begin{table}[!b]
  \caption{\footnotesize{Ablation study of different model components. We report top-1 predicted accuracy (\%) results of different evaluation scenarios for a ViT-T model on the first trial of ROD. The best results are highlighted in bold.}}
  \label{tab:ablations}
  \resizebox{\linewidth}{!}{%
  \centering
  \begin{tabular}{lccc}
    \toprule
    \multicolumn{1}{l}{\textbf{Method}} &
    \multicolumn{1}{c}{\textbf{k-NN}}  &
    \multicolumn{1}{c}{\textbf{Lin.Eval}} &
    \multicolumn{1}{c}{\textbf{Finetune}} \\
    \midrule
    RGB  & $79.5$ & $80.2$ & $83.0$ \\
    \midrule
    Depth (Raw)   & $59.7$ & $62.0$ & $69.9$ \\
    Depth (HHA) & $65.1$ & $67.8$ & $73.2$ \\
    Depth (SurfNorm)  & $66.3$ & $69.9$ & $76.7$ \\
    \midrule
    RGB-D (Early w/ dual-emb)   & - & - & $82.1$ \\
    RGB-D (Early w/ joint-emb.)   & - & - & $81.0$ \\
        \midrule
    RGB-D (Late w/ avg)  & $82.0$ & $82.1$ & $85.5$ \\
    RGB-D (Late w/ max)  & $\mathbf{85.4}$ & $85.7$ & $88.4$ \\
    RGB-D (Late w/ cat)  & $\mathbf{85.4}$ & $\mathbf{87.4}$ & $\mathbf{90.0}$ \\
    \bottomrule
  \end{tabular}}
\end{table}

We ablate the following aspects of our approach: a) the format of the input depth image, as described in Sec.~\ref{depth_rep}, b) the type of RGB-D fusion used (\textit{Early} vs. \textit{Late}), c) the type of embedder used in the \textit{Early} fusion baseline (\textit{joint-} vs. \textit{dual-emb.}) as well as d) the type of fusion used in the \textit{Late} fusion baseline.
We experiment with ViT-T using only the first trial of ROD's evaluation setup.
In order to gain better insight into all different components, we use three different evaluation setups, namely: a) k-nearest neighbor (\textit{k-NN}) on top of frozen embeddings, b) training a linear head on top of frozen embeddings (\textit{Lin.Eval}), and c) fine-tuning the model end-to-end with an MLP head for classification (\textit{Finetune}). 
We note that the early baseline does not include results with frozen embeddings, as ViT cannot be used out-of-the-box for depth embeddings.
For RGB-D methods we report results using the \textit{SurfNorm} depth format, as it achieves the best results.
For \textit{k-NN}, we experimentally find that $k=3$ and cosine similarity as the distance metric yields best results. 
For \textit{Lin.Eval}, we train with SGD with momentum value $0.9$, batch size $128$ and a learning rate of $5 \cdot 10^{-4}$. 
For fine-tuning, we use AdamW \cite{adamW}, batch size $512$, a learning rate of $9 \cdot 10^{-5}$ and a linear decay over 10 epochs. We train on 2 Nvidia Titan Xp GPUs. 
Our results are summarized in Table~\ref{tab:ablations}. 

We observe that the \textit{SurfNorm} depth representation leads to the best depth-only performance in ViT, which is in compliance with previous works.
Regarding early fusion, we also find that using separate embeddings for the two modalities (\textit{dual} baseline) leads to marginally better results than the \textit{joint}. 
Regarding late fusion, the concatenation operation outperforms other fusion approaches, with the cost of doubling the classifier's hidden size.
When comparing the two fusion methods, we observe that the late fusion baseline far outperforms the early one. 
In particular, the early baseline achieves worst results than RGB-only. 
We believe that this result reinforces our original hypothesis, namely that in low-data RGB $\rightarrow$ RGB-D transfer scenarios, attempting to modify the input RGB-D embedder leads to overfitting, as the depth embedding layer has to be trained from scratch. 
In contrast, in the late fusion baseline, the encoder uses the same embedding weights but only learns to adapt the final layers to incorporate depth features. 
We confirm this insight by verifying that the weights in the adapted ViT have a greater average absolute difference in the early rather than the late baseline.
\subsection{Offline RGB-D Object Recognition}
\label{wrgbd}

Table~\ref{tab:wrgbd} presents results for the $10$ trials of the ROD evaluation setup, compared with previous state-of-the-art methods, as reported in \cite{sota10}.
We use ViT-B and the best configuration from our ablation experiments (i.e. \textit{Late} + \textit{cat}). 
For reference, we include results over the 10 trials using the dual-embedder early fusion architecture.
We also include another baseline, ViT-B Ensemble (\textit{Ens.}), in which we ensemble separately fine-tuned models for RGB and depth. 
Finally, we further scale our model using the ViT-L and SWIN-B \cite{swin} models, in order to explore the performance gains when adding model capacity.
We train using the same hyper-parameters as the fine-tuning experiments of the previous section, but with a learning rate of $3 \cdot 10^{-5}$ and batch size of $64$.

\begin{table}[!t]
  \caption{\footnotesize{Top-1 accuracy (\%) over the 10 trials of ROD. The best results are highlighted in bold, and the second-best results are underlined. The method with $^{\dagger}$ uses one encoder per modality, thus doubling the spatial requirements.}}
  \label{tab:wrgbd}
  \resizebox{\linewidth}{!}{%
  \centering
  \begin{tabular}{lccc}
    \toprule
    \multicolumn{1}{l}{\textbf{Method}} &
    \multicolumn{1}{c}{\textbf{RGB}}  &
    \multicolumn{1}{c}{\textbf{Depth}} &
    \multicolumn{1}{c}{\textbf{RGB-D}} \\
    \midrule
    Fusion 2D/3D CNNs \cite{sota1} & $89.0 \pm 2.1$ & $78.4 \pm 2.4$ & $91.8 \pm 0.9$ \\
    STEM-CaRFs  \cite{sota2} & $88.0 \pm 2.0$ & $80.8 \pm 2.1$ & $92.2 \pm 1.3$ \\
    MM-LRF-ELM \cite{sota3} & $84.3 \pm 3.2$ & $82.9 \pm 2.5$ & $89.6 \pm 2.5$ \\
    VGG f-RNN \cite{sota4} & $89.9 \pm 1.6$ & $84.0 \pm 1.8$ & $92.5 \pm 1.2$ \\
    DECO \cite{sota5}  & $89.5 \pm 1.6$ & $84.0 \pm 2.3$ & $93.6 \pm 0.9$ \\
    MDSI-CNN \cite{sota6} & $89.9 \pm 1.8$ & $84.9 \pm 1.7$ & $92.8 \pm 1.2$ \\
    HP-CNN \cite{sota7} & $87.6 \pm 2.2$ & $\underline{85.0 \pm 2.1}$ & $91.1 \pm 1.4$ \\
    RCFusion \cite{sota8} & $89.6 \pm 2.2$ & $\underline{85.9 \pm 2.7}$ & $\underline{94.4 \pm 1.4}$ \\
    MMFLAN \cite{sota9} & $83.9 \pm 2.2$ & $84.0  \pm 2.6$ & $93.1 \pm 1.3$ \\
    DenseNet121-RNN \cite{sota10} & $91.5 \pm 1.1$ & $86.9 \pm 2.1$ & $93.5 \pm 1.0$ \\
    ResNet101-RNN \cite{sota10}  & $\underline{92.3 \pm 1.0}$ & $\mathbf{87.2 \pm 2.5}$ & $94.1 \pm 1.0$ \\
    \midrule
    Ours (ViT-B Ens.)$^{\dagger}$ & $90.8 \pm 1.9$ &  $83.7 \pm 2.1$  & $90.4 \pm 1.5$ \\
    Ours (ViT-B Early)  & - & - & $89.5 \pm 1.5$ \\
    Ours (ViT-B Late) &  $\underline{92.6 \pm 1.1}$  & $83.6 \pm 2.4$ & $\underline{94.8 \pm 1.5}$ \\
    Ours (ViT-L Late) & $\underline{92.9 \pm 1.3}$ & $83.5 \pm 2.1$ & $\underline{95.1 \pm 1.3}$ \\
    Ours (SWIN-B Late) & $\mathbf{93.1 \pm 1.6}$ & $\underline{85.0 \pm 2.3}$ & $\mathbf{95.4 \pm 1.3}$ \\
    \bottomrule
  \end{tabular}}
  \vspace{-3mm}
\end{table}

\begin{table}[!b]
\centering
\vspace{-1.5mm}
  \caption{\footnotesize{Top-1 predicted accuracy (\%) of 10-fold cross validation in various RGB-D datasets. Dataset with $^{\dagger}$ uses surface normals instead of raw depth maps. Best results are highlighted in bold. }
  }
  \label{tab:more_offline}
  \resizebox{\linewidth}{!}{%
  \centering
  \begin{tabular}{lccc}
    \toprule
     \textbf{Method} & \textbf{ARID40k} & \textbf{HB}$^{\dagger}$ & \textbf{OCID}  \\
        \midrule
        ViT-B (RGB) & $97.8 \pm 0.2$ & $98.1 \pm 0.2$ & $91.7 \pm 1.5$  \\
        ViT-B (Depth) & $51.1 \pm 1.0$ & $82.9 \pm 0.8$ & $52.2 \pm 2.5$ \\
        ViT-B (RGB-D \textit{Early}) & $96.5 \pm 0.4$ & $97.9 \pm 0.3$ & $84.9 \pm 2.2$ \\
        ViT-B (RGB-D \textit{Late}) & $\mathbf{98.3 \pm 0.2}$ & $\mathbf{99.6 \pm 0.2}$ & $\mathbf{93.5 \pm 1.4}$  \\
    \bottomrule
  \end{tabular}
  }
\end{table}

We observe that our approach achieves new state-of-the-art in the ROD benchmark, even with the baseline ViT-B model,.
When scaling to larger models, our approach achieves a delta of $1.0\%$ from the previous best result in RGB-D. 
Compared to our ensemble baseline, we observe that joint fine-tuning indeed leads to better scores than ensembling two modality-specific encoders, as the latter fails to model cross-modal interactions during training, instead fusing unimodal features only for inference.

We further conduct 10-fold cross validation experiments in the remaining RGB-D datasets and report results in Table~\ref{tab:more_offline}.
In the absence of benchmark scores by previous works, we only evaluate ViT-B in the different modality and fusion settings.
We train for $4$ epochs with same hyper-parameters, except a batch size of $16$ for OCID.
Similarly to ROD, \textit{SurfNorm} is the most effective depth format, and the late fusion baseline achieves best results, with the gap to early fusion increasing as the training data decreases ($8.6\%$ in OCID with $5.5k$ samples, $1.8\%$ for ARID with $40k$ samples).

\subsection{Synthetic-to-Real Transfer for RGB-D Object Recognition}
\label{synthetic-to-real}
In this section, we explore whether our recipe for fine-tuning Transformers in RGB-D can aid in visual domain adaptation.
To test in a synthetic-to-real adaptation scenario, we train a ViT-B in the synthetic version of ROD, the SynROD dataset \cite{SynROD}, which recreates the objects of ROD in Blender and selects an equal amount of examples per category and instance to generate an equal-sized counterpart. 
We train ViT-B in SynROD using the same hyperparameters of the previous section and compare with the performance of training in real data only, after fine-tuning for different volumes of real examples ($1,5,10,20$).
We use the first trial of ROD as a train-test split and experiment with RGB-only, Depth-only, and both early and late fusion RGB-D baselines.
Results are shown in Table~\ref{tab:syn2real}.

\begin{table}[!t]
  \caption{\footnotesize{Top-1 accuracy (\%) for transferring ViT-B baselines from synthetic (SynROD) to real (ROD) RGB-D images, using different amounts of fine-tuning data per object instance (1, 5, 10, 20). The best results are highlighted in bold.}}
  \label{tab:syn2real}
  \resizebox{\linewidth}{!}{%
  \centering
  \begin{tabular}{lccccc}
    \toprule
    \multirow{2}{2.1em}{\textbf{Setup}} & 
    \multirow{2}{3em}{\textbf{\#Data}} & 
    \multirow{2}{2.1em}{\textbf{RGB}} & 
    \multirow{2}{2.1em}{\textbf{Depth}} & 
    \multicolumn{2}{c}{\textbf{RGB-D}} \\
    \cmidrule(l{8pt}r{10pt}){5-6}
     & & & & Early & Late \\
    \midrule
    ROD  - Only  & - & $\mathbf{90.6}$  & $\textbf{81.9}$ & $89.3$ & $92.7$ \\
    \midrule
    SynROD - Only & $0$ & $44.2$ & $9.1$ & $40.1$ & $44.7$\\
    SynROD$\rightarrow$ROD & $1$ & $75.8$  & $44.8$ & $67.7$ & $76.1$  \\
    SynROD$\rightarrow$ROD & $5$ & $88.4$ & $61.0$  & $80.5$ & $90.8$ \\
    SynROD$\rightarrow$ROD & $10$ & $89.7$ & $67.9$ & $85.1$ & $92.7$\\   
    SynROD$\rightarrow$ROD & $20$ & $90.5$ & $73.0$ & $86.2$ & $\textbf{93.7}$\\   
        \bottomrule
  \end{tabular}}
\end{table}
We observe that as in the offline setting, unimodal baselines struggle to classify all object categories without utilizing both modalities, with depth having a dramatic margin due to the distribution gap between smooth synthetic and noisy real-depth images.
The early fusion baseline struggles to model cross-modal interactions with only a few data.
In contrary, our late fusion baseline is the only one that surpasses the performance of the model trained in real data while using only 20 labeled examples per object instance. 
On-par performance can be achieved with only 10 examples.
We believe that this result encourages the use of late fusion in synthetic-to-real RGB-D domain adaptation scenarios.

\subsection{Open-Ended Lifelong RGB-D Object Recognition}
\label{lifelong}

\begin{figure}[!t]
    \centering
    \includegraphics[width=\columnwidth]{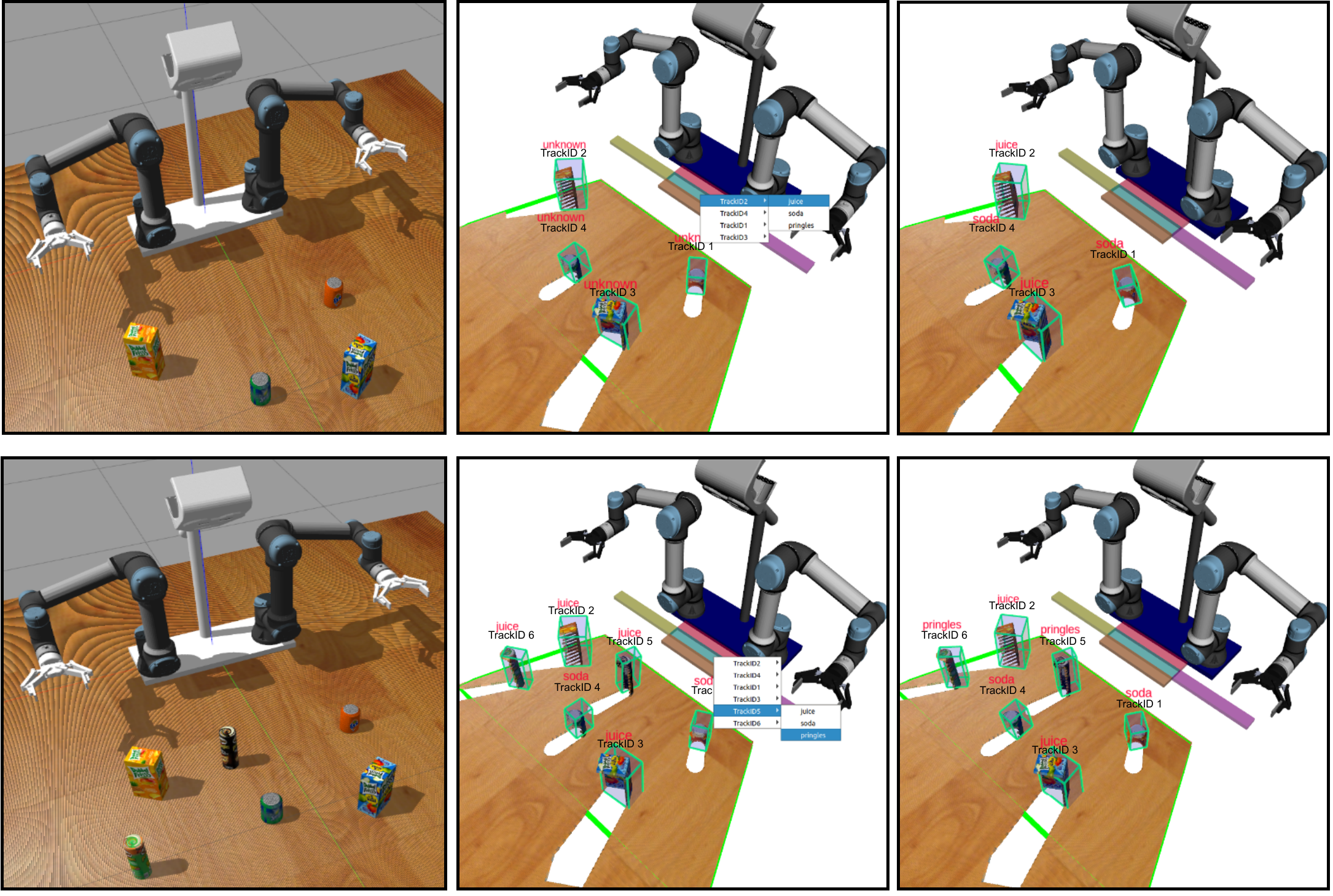}
    \caption{ \footnotesize{We implement an open-ended lifelong object recognition pipeline based on our ViT-\textit{late} fusion model. The user teaches a new category by labeling a tracked object in a GUI \textit{(top-middle)} and the system correctly recognizes all different instances of that category \textit{(top-right)}. The user can introduce new categories for new scenes \textit{(bottom-left)} or correct mistakes from the system \textit{(bottom-middle)}. The robot adapts in a lifelong fashion to recognize all objects correctly \textit{(bottom-right)}. \textit{(Best viewed with zoom)}.}}
    \label{fig:sim_user}
    \vspace{-5mm}
\end{figure}

In this section, we evaluate our approach in an online fashion, where we assume that the learning agent is presented with novel object instances throughout a lifespan.
We follow the evaluation protocol proposed in \cite{Kasaei2019OrthographicNetAD,LocalLDA, GOOD}.
In particular, we develop a simulated user who gradually introduces new object categories by presenting an unseen view of a novel object (see Fig.~\ref{fig:sim_user}). 
Teaching the agent is performed until a specific \textit{protocol threshold} value is met (e.g for value $0.67$, accuracy rate must be at least double from error rate), after when a new category is introduced, or the agent learned all existing categories. 
We note that random sampling is used to select novel views from each category.

\begin{table}[!b]
  \vspace{-5mm}
  \caption{\footnotesize{Online lifelong learning evaluation in ROD with a simulated teacher. Accuracy results for previous works were reported without decimals. The arrow demonstrates if better results are higher or lower for each metric (\textit{see text for explanation of the evaluation metrics).}.}
  }
  \label{tab:online}
  \resizebox{\linewidth}{!}{%
  \centering
  \begin{tabular}{lccccc}
    \toprule
     \textbf{Method} & \textbf{QCI}$\mathbf{\downarrow}$ & \textbf{ALC}$\mathbf{\uparrow}$ & \textbf{AIC}$\mathbf{\downarrow}$ & \textbf{GCA}$\mathbf{\uparrow}$ & \textbf{APA}$\mathbf{\uparrow}$  \\
        \midrule
        RACE \cite{Oliveira20163DOP} & $382.1$ & $19.9$ & $8.9$ & $67.0$ & $78.0$ \\
        BoW \cite{kasaei2018towards} & $411.8$ & $21.8$ & $8.2$ & $71.0$ & $82.0$ \\
        Open-Ended LDA \cite{Hoffman2010OnlineLF} & $\mathbf{262.6}$ & $14.4$ & $9.1$ & $66.0$ & $80.0$ \\
        Local LDA \cite{LocalLDA} & $613.0$ & $28.5$ & $9.1$ & $71.0$ & $80.0$ \\
        GOOD \cite{GOOD} & $1659.2$ & $39.2$ & $17.3$ & $66.0$ & $74.0$ \\
        OrthographicNet \cite{Kasaei2019OrthographicNetAD} & $1342.6$ & $\mathbf{51.0}$ & $10.0$ & $77.0$ & $80.0$ \\
                \midrule
        Ours (ViT-B \textit{Late}) & $1327.0$ & $\mathbf{51.0}$ & $\mathbf{5.3}$ & $\mathbf{86.4}$ & $\mathbf{88.7}$ \\
             \bottomrule
  \end{tabular}}
\end{table}

In order to measure the effect of \textit{catastrophic forgetting}, the user tests the agent in all previous categories after each new object introduction.
The evaluation stops either when the agent has learned all categories (according to the protocol threshold) or is unable to do after exceeding a specified number of \textit{Question/Correction Iterations (QCI)}. 
Evaluation metrics include: \textit{(i)}, \textit{Average number of Learned Categories (ALC)}, \textit{(ii)}, \textit{Average number of stored Instances per Category (AIC)}, \textit{(iii)}, \textit{Global Classification Accuracy (GCA)} and \textit{(iv)}, \textit{Average Protocol Accuracy (APA)}. 
We apply our protocol in the ROD dataset and compare it with published scores from \cite{Kasaei2019OrthographicNetAD}, averaging results from 10 individual runs with a protocol threshold of $0.67$.
Similar to baselines, we use our model as a feature extractor, paired with a k-NN classifier with an expandable support set (i.e., novel views introduced by the human teacher).
Results are shown in Table~\ref{tab:online}.
Our approach far exceeds the performance of previous works across metrics, showcasing the representational strength of ViTs when transferred to RGB-D scenarios. 
In particular, our ViT model is able to learn all $51$ categories using $\sim 5$ instances per category and achieves accuracy metrics with a delta of $>8\%$ compared to the previous best result.

\begin{figure*}[!t]
    \centering
    \includegraphics[width=1\textwidth]{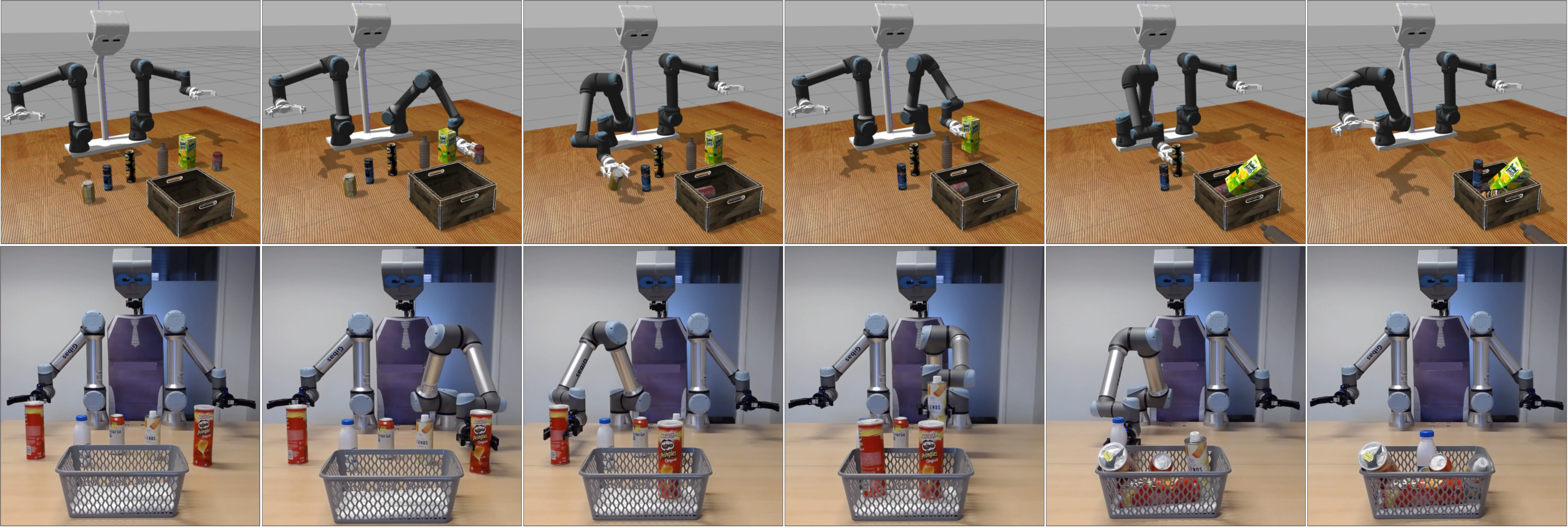}
    \caption{\footnotesize{A sequence of snapshots capturing the experimental setup and the behavior of the robot in Gazebo (\textit{top)} and in a real-world (\textit{bottom)}. We randomly place objects and instruct the robot to recognize them and place them in the container.}}
    \label{fig:robot_setup}
\end{figure*}

Furthermore, we conduct ablation studies to verify whether the attested performance comes indeed from the RGB-D fusion technique, or if it is inherited from ImageNet pretraining, i.e. an RGB-only model would suffice. 
To that end, we compare our late fusion model with unimodal baselines.
We also include experiments with stricter teachers, by setting the protocol threshold value to $\{0.7, 0.8, 0.9\}$.
As before, we use $k=3$ and the cosine distance function and perform one run per configuration.
Results are summarized in Table~\ref{tab:protocol}.
We observe that even without any RGB-D fine-tuning, fusing the embeddings generated by the ImageNet ViT checkpoint still provides accuracy benefits over unimodal embeddings, in all protocol threshold settings.

\begin{table}[!b]
  \caption{\footnotesize{Ablation studies for online lifelong learning evaluation in ROD with a ViT-B model. Higher protocol threshold values represent a "stricter" teacher, requiring more correct answers to consider that the agent learned a category successfully. We report results for RGB-only, Depth-only, and our RGB-D late fusion baseline.}
  }
  \label{tab:protocol}
  \resizebox{\linewidth}{!}{%
  \centering
  \begin{tabular}{clccccc}
    \toprule
    \multicolumn{1}{c}{\textbf{Threshold}} &
    \multicolumn{1}{l}{\textbf{Method}} &
    \multicolumn{5}{c}{\textbf{Washington RGB-D Objects}} \\
    \cmidrule(l{8pt}r{10pt}){3-7}
     & & \textbf{QCI}$\mathbf{\downarrow}$ & \textbf{ALC}$\mathbf{\uparrow}$ & \textbf{AIC}$\mathbf{\downarrow}$ & \textbf{GCA}$\mathbf{\uparrow}$ & \textbf{APA}$\mathbf{\uparrow}$  \\
    \midrule
     & ViT-B (RGB)  & $1325$ & $\mathbf{51}$ & $6.4$ & $86.9$	&	$88.2$   \\
    $\mathbf{0.7}$ & ViT-B (Depth)  & $1329$ & $\mathbf{51}$ & $7.7$ & $77.7$	&	$79.2$   \\
     & ViT-B (RGB-D \textit{Late})  & $1325$ & $\mathbf{51}$ & $\mathbf{5.9}$ & $\mathbf{88.3}$	&	$\mathbf{89.1}$ \\
     \midrule
     & ViT-B (RGB) & $1369$ & $\mathbf{51}$ & $\mathbf{6.1}$ & $88.6$	&	$89.4$  \\
    $\mathbf{0.8}$ & ViT-B (Depth) & $2029$ & $\mathbf{51}$ & $7.8$ & $81.8$	&	$84.2$   \\
     &ViT-B (RGB-D \textit{Late}) & $1370$ & $\mathbf{51}$	&	$6.3$ &	$\mathbf{88.7}$	&	$\mathbf{90.2}$    \\
    \midrule
     & ViT-B (RGB) & $2368$ & $\mathbf{51}$ & $7.3$ & $90.2$	&	$93.1$ \\
    $\mathbf{0.9}$ & ViT-B (Depth) & $2954$ &  $34$ & $8.4$  & $90.1$ & $90.5$  \\
     & ViT-B (RGB-D \textit{Late}) & $1695$ & $\mathbf{51}$ & $\mathbf{6.7}$ & $\mathbf{90.7}$	&	$\mathbf{93.9}$  \\
     \bottomrule
  \end{tabular}}
\end{table} 
\subsection{Robot Demonstrations}
\label{robot}
We develop a simulation environment in Gazebo to evaluate the real-time performance of the proposed approach in the context of a \texttt{clean\_table} task (see Fig.~\ref{fig:robot_setup}). 
For this round of experiments, we integrate our work into the robotic system presented in~\cite{kasaei2018towards}. 
At the beginning of each experiment, we randomly place four to six objects and a container on the table. 
The robot doesn't have any knowledge about the objects, therefore, it recognizes all objects as ``\textit{unknown}''. 
A human user teaches new object categories to the robot using a GUI and the robot recognizes all object instances before placing them into the container.
Note that the pose of the container is known to the robot in advance. 
We performed $10$ experiments to validate the performance of the robot.
In all experiments, we observed that the robot could incrementally recognize all object categories using a single instance for teaching, and completed the task successfully.
A video of these experiments has been attached to the paper as supplementary material. 

\section{Conclusion}
In this work, we propose a simple yet strong recipe for transferring pretrained ViTs in RGB-D domains. 
We demonstrate that, unlike most prior state-of-the-art that use early fusion, the late fusion strategy transfers better in the low-data regime. 
By fine-tuning a ViT with our late fusion approach, we push the state-of-the-art in the Washington RGB-D Objects benchmark by 1.0\% and find that compared to unimodal approaches, late RGB-D fusion aids in few-shot visual domain adaptation.
We further illustrate that our approach outperforms previous works when the training-test paradigm is replaced with an open-ended lifelong learning scenario and demonstrate its utility for a robotic task, both in simulation and with a real robot.
We hope that our approach will lead to more research on efficiently transferring ViTs for robotics-specific domains.
This work leaves us with a multitude of potential future directions, regarding the sophistication of the RGB-D fusion, computational efficiency and generalization to novel domains.
In the future, we plan to experiment with hierarchical instead of pooling-based fusion, adapters for parameter-efficient fine-tuning and comparing supervised vs. unsupervised pretraining for transfer in domains outside the ImageNet class distribution.


\printbibliography

\end{document}